# Boosting classification reliability of NLP transformer models in the long run

**Zoltán Kmetty*[1,2], Bence Kollányi[1], Krisztián Boros[1]**

[1]*Centre for Social Sciences, CSS-RECENS Research Group, Budapest*
[2]*Eötvös Loránd University, Faculty of Social Sciences, Sociology Department, Budapest*

*Corresponding author*

*Zoltán Kmetty, kmetty.zoltan@tk.hu. Centre for Social Sciences, CSS-RECENS Research Group, 1097 Budapest, Tóth Kálmán u. 4., Hungary*

**Introduction**

A key goal of machine learning projects is some form of classification of the input data. This classification is typically done so that both the training and the data to be classified come from the same period and data set. In practice, however, it may often be the case that a particular classification is extended to a different set of data and/or to a different period. The need and the possibility to extend classification over time is strongly supported by increasing digitization as updated datasets are more frequently available for industrial and scientific research purposes. But how long is a classification suitable for, and when is it worth re-training our model? Since the much-quoted Google Flu case, every researcher using machine learning knows that models should not be blindly trusted and that it is crucial to revise them in time (Lazer et al., 2014). The need for retraining may be particularly relevant in cases where the domain under study changes rapidly over time. In the various natural language processing (NLP) projects, this problem is common, as language and language use can change within a short period of time on a given topic (Kulkarni et al., 2015). Furthermore, neural network-based black-box models complicate the problem, as they offer little insight into how a particular classification model works, making it harder to identify what changes in content or context might cause the model to break down. This black-box feature is a problem with the transformer-based NLP models currently used for classification tasks in data-mining projects.

Transformer-based machine learning models have become an important tool for many natural language processing (NLP) tasks since the introduction of the method (Vaswani et al., 2017). Connecting the encoder and decoder through an attention mechanism (which can capture the meaning of an input sequence, e.g., a sentence) and effective parallelization of the training process led to robust generalized language models. Transformer models are usually pre-trained on a generic language corpus and can later be tuned to domain-specific tasks. These models can be used for a wide array of general NLP tasks, from classification problems to document summarization, sentiment analysis, and named entity recognition.

When a model is trained and fine-tuned on a range of training materials, it is often difficult to apply it to another field (Ramponi and Plank, 2020). A growing body of literature suggests that domain shifts affect the performance of general models. Performance can significantly drop when a model is applied to a dataset different from the data used for pre-training, often referred to as an out-of-distribution (OOD) dataset (Koh et al., 2021; Larson et al., 2021; Ma et al., 2019; Wang et al., 2022).

Language models trained on general texts are often insufficient for tasks related to special domains, such as medical texts or legal documents. The performance of Bidirectional Encoder Representations from Transformers (BERT) can also degrade when the general model is applied to a specific domain, so further fine-tuning is often required for specific NLP tasks. BERT, one of the most popular transformer-based models, was originally pre-trained on the complete Wikipedia corpus and books digitized by the Google Books project (Devlin et al., 2018). After BERT was released as open source in 2019, researchers focusing on specific domains created their own specialized versions of BERT by extending the base model and pre-training their models on domain-specific text. For example, BioBERT was trained on biomedical corpora (PubMed abstracts and full-text articles) to learn the biomedical language (Lee et al., 2019), SciBERT was pre-trained on a random sample of 1.14 million scientific (mainly computer science and biomedical) articles (Beltagy et al., 2019), while FinBERT was pre-trained on a large financial corpus consisting of more than 46,000 news articles with 29 million words (Araci, 2019). These domain-specific models can be fine-tuned to downstream NLP tasks with great success and significantly improve the performance of BERT on these domain-specific tasks. Using a large domain-specific corpus and pre-training the model with it is often difficult and expensive. Another approach is to pre-train the model using the training classification dataset, which is often a much smaller corpus (Araci, 2019).

In addition to the difficulties of applying a language model to an OOD dataset (a domain distinct from the materials used to train the model), the performance of trained language models may also degrade over time. Measuring the robustness of such a model over time has recently attracted considerable research attention. Measuring the performance of transfer models is challenging because transformer models tend to overestimate their performance when the training and evaluation periods overlap (Lazaridou et al., 2021). The literature suggests temporal misalignment degrades performance when the testing period is temporally distant from the training period (Luu et al., 2022; Röttger and Pierrehumbert, 2021).

Previous work also examined the changes in the language itself over time. The pace of language change has become increasingly rapid in the context of social media (Goel et al., 2016). Linguistic shifts can also affect the performance of classification tasks based on machine learning. Both changes in word usage and altering the context of the same words can lead to performance degradation (Huang and Paul, 2019). Temporal shifts have been studied in the context of several NLP tasks, including domain classification (Huang and Paul, 2018; Agarwal & Nenkova, 2022), named entity recognition (NER) (Rijhwani and Preotiuc-Pietro, 2020; Chen et al., 2021), and sentiment analysis (Lukes and Søgaard, 2018).

Florio et al. (2020) evaluated the temporal robustness of an Italian version of the model BERT and found that the model's efficiency is highly time-dependent. Although the Italian BERT model was trained on social media data and it was fine-tuned on a labeled dataset on hate speech, the content of the texts used for testing changed rapidly. When the temporal distance between the training set and the test set increased, the model's precision and F1 values decreased significantly. The authors concluded that the language of hateful content is closely linked to current events in the news that trigger hate speech and is therefore highly sensitive to changes over time.

Florio et al. also found that training the model repeatedly with new data significantly helps the model to maintain its performance over time. The authors compared two different solutions to this problem. First, they used a sliding window approach in which the model was fine-tuned on a data set that was collected one month prior to the test data set. The other solution used an incremental approach, meaning that the model was trained with data from all available data before the test period so that after a longer period, data from several months were used. BERT performed significantly better in their experiment than a Support Vector Machine (SVM) model. It proved to be more robust and less sensitive to changes over time, even when fine-tuned on a static dataset and not adjusted by using more recent labeled data.

We answer two research questions in this paper. On the one hand, we analyze the best approaches to fine-tune a BERT model on a long-running classification task regarding the temporal selection of fine-tuning data (RQ1). On the other hand, we test a particular re-classification approach where we use comments selected by their temporal uniqueness (RQ2).

**Data**

This paper uses comments about COVID-19-related vaccination in Hungary collected between September 2020 and December 2021. We used a social listening platform called SentiOne to collect the data. SentiOne (www.sentione.com) is an international social listening software, a content-based web analytics platform that covers and recognizes 30 languages across Europe. It collects, indexes, and analyses online public content from social media platforms, blogs, forums, and websites. By selecting keywords and tags used by the users, the interactive platform returns content that is mentioned either in the text or in its metadata. We set the start date to September 2020, as this is when the debate around Covid-19 vaccines started in Hungary. For creating our corpus, we used keywords related to vaccination. Then, we filtered out stop-words and used topic-modeling approaches to find the relevant content. There were several articles that were not related to the topic in the initial corpus, for example, articles related to animal vaccination. We retained only those articles and posts that relate to Covid-19 vaccination. The cleaned corpus consisted of about 295,152 articles and posts on social media. In this paper, we do not use the articles themselves, just the comments under the articles and relevant social media posts. At the comment level, we retained those comments written under the selected articles or containing vaccination or vaccine-related keywords. The corpus contains 8,260,808 comments from the selected period.

One of our initial project goals was to find the anti-vaccination content in the comments. To this end, we developed a transformer-based classifier in early 2021. As the first step, the researchers in the project coded a random sample of 1,000 comments to anti-vax and non-anti-vax or not related to vaccination category. Based on this initial classification, we defined a set of keywords that covered well (over 98 percent recall) the anti-vax comments (see the supplementary for the list of keywords). In this step, we wanted to remove those comments from the classification where we did not expect to find any anti-vax content. This method filtered out 27.5 percent of the comments.

After filtering the comments with the keywords, we randomly selected 10,000 comments from the pool of 1.7 million. The selected comments covered the period from September 2020 to February 7, 2021. We double-annotated these comments to anti-vaxxer comments and pro-vaxxer or neutral comments. In case of disagreement between annotators, a supervisor selected the final category. The supplementary contains the details of the annotation process. The kappa value for the inter-coder agreement was 0.73.

These manually classified 10,000 comments formed the basis for categorizing the rest of the comment pool. For the comment categorization, we used a transformer-based NLP model based on the Hungarian version of BERT (huBERT) (Nemeskey, 2021). The average macro-precision of the classification was 0.79, and the recall and F1 scores were 0.78. This model was not fine-tuned to the topic. We call this model the base model in this paper.

We continued to collect COVID-related content in the project on a daily basis and classified the new comments using the above-described transformer model. In the last months of 2021, we decided to run a second round of classifications to check whether our BERT classifier could still classify the comments with high validity. In this round, we selected comments from November 2021. To overcome the unbalanced ratio of anti-vax and non-anti-vax comments, we increased the weighting of the anti-vax comments in this annotation round. Before annotation, we classified the comments using our BERT classifier and selected 3,000 anti-vax and 1,500 pro-vaxxer or neutral comments. The 4,500 comments were annotated in the same way as in the first round of annotation. Two annotators annotated all the comments. The annotators did not know how the BERT classifier classified the comments. If there was a difference between the annotators, a supervisor selected the final category. The kappa value for the inter-coder agreement was 0.63. In this paper, we use some of these annotated comments from the second round of annotation to test the quality of the different classification approaches (see the details in the method section).

**Methods**

Our primary research question (RQ1) was how we could optimize the classification of vaccination-related comments in the long run. We use data from different periods in the first set of tests to fine-tune the original BERT model. The classification logic was the following. We started with the original huBERT model; then, we fine-tuned the model with non-labeled data. After the fine-tuning, we add the labeled data from the first annotation round to the model to learn the classification. The quality of this classification was measured on two bases. We leave out 30 percent of the comments from the labeled training set for test purposes. These comments cover the period from September 2020 to January 2021. The second quality check was on the test comments selected from the second annotation round. These comments are coming from November 2021. For the fine-tuning, we use non-labeled data from three different periods:

- 09/2020-01/2021 – 1.3m comments
- 10/2021-11/2021 – 777k comments
- 09/2020-12/2021 – 5.5m comments

Again, we tested a different approach. We used unlabeled data from the entire period (09/2020-12/2021) for fine-tuning, but we only allowed two epochs for the model to specify the weights; then we used these model weights to do further fine-tuning with data from October and November 2021. Here, we wanted to test whether it is worth combining shallow and deep learning techniques to maximize classification quality.

In addition to testing the different fine-tuning approaches, we also measured how a second round of annotation could boost the quality of the classification (RQ2).

We tested an optimization approach for the selection of second-round annotated comments. To do this, we computed a weirdness statistic for all words in the comments. Weirdness is a simple statistic on a word level, which shows the ratio of a given word between two corpora or two time-periods (Basile, 2020). In our case, we calculated the baseline ratio of a word for the period between September 2020 and January 2021. This is a simple division of the frequency of the word and the total counts of all words in the selected period. We then calculated a weirdness statistic for words on a monthly basis. We calculated the ratio of one word compared to all words in the selected month and divided this number by the baseline ratio. This weirdness number is one if the occurrence of a word in a month is the same as in the baseline period, and it rises above one if the words appear more often in that month. To compute the weirdness statistics, we used a sample of comments. We selected 40,000 comments each month and lemmatized the texts with a Hungarian lemmatizer e-magyar (Zsibrita, 2013). We used the lemmatized comments for the calculation.

Figure 1.A presents the weirdness value of the word "Delta" and Figure 1.B the Google searches for this word. WHO began using the delta-variant name in June 2021. The weirdness statistic for this word goes parallel with the number of Google searches, which started to increase in the same month and peaked in August 2021.

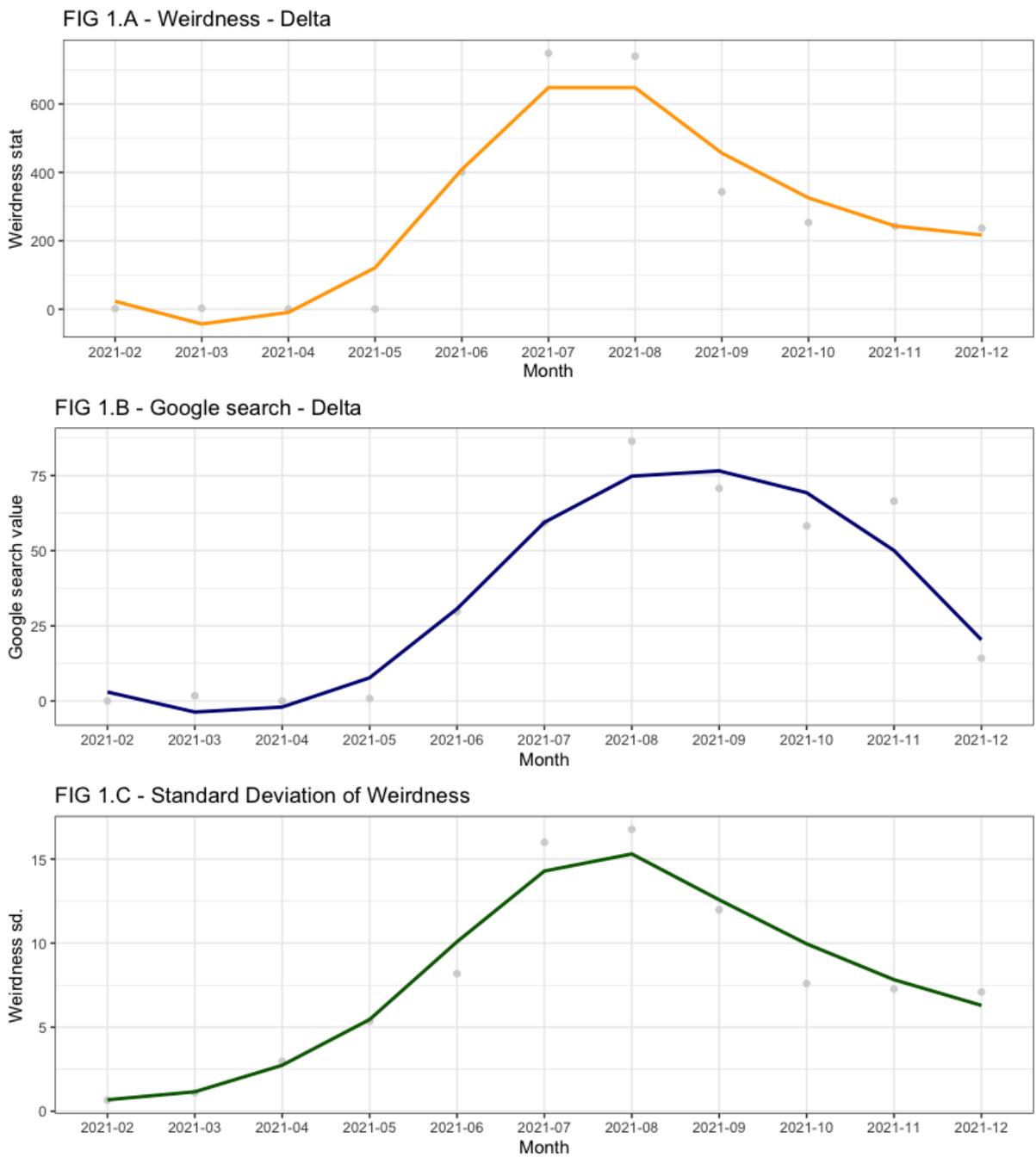

*Figure 1. A: Weirdness value for the word Delta. B: Google searches in Hungary for the word Delta. C: The monthly standard deviation of the weirdness statistics*

Weirdness is a perfect statistic to measure the change in our corpus compared to a previous period. Figure 1.C shows the change in the standard deviation of the weirdness statistic. A standard deviation of zero would mean that the word ratio in the observed month is the same as in the baseline period. As expected, the standard deviation of the weirdness statistic was the lowest in February 2021, the month closest to the baseline period. The standard deviation value increased until August 2021, and then decreased. Seasonality might explain this decline in word usage.

We calculated a comment level weirdness based on the word level weirdness statistic. It is a simple average of the word's monthly weirdness value per comment. We use this comment level weirdness value to select a particular sample from the second round of annotated contents by using weirdness statistics as a weight in the sampling process. We selected 1,500 comments based on this approach and 1,500 comments randomly from the pool of secondly annotated comments. There was some overlap (693 comments) between the weirdness based 1,500 comments and the randomly selected 1,500 comments. Both samples were selected from the second annotation round of annotation, covering November 2021. The 2,193 comments from the second annotation round used for test purposes did not overlap with these samples. We used the two samples of annotated comments to boost the classification. We added these comments to the original training dataset and trained the models with these extended training sets. In these tests, we used all the available unlabeled data for fine-tuning.

**Results**

We first present the model performance for the first period of annotation. Of the 10,000 comments annotated in the first round, we withheld 30 percent for testing and trained the models on the remaining comments. This round of annotation covered the period between September 2020 and January 2021. Figure 2. summarizes the results. The base huBERT model, without any fine-tuning, achieved an F1-score of 0.65 in the anti-vaxxer category and 0.92 for the non-anti-vaxxer and neutral comments. As expected, fine-tuning boosted the performance of our transformer model. We obtained an F1-score of 0.73 for the anti-vaxxer comments when we used the complete unlabeled data for fine-tuning. We obtained the same value when we fine-tuned with data selected from the same period as the test comments, namely September 2020 to January 2021. As expected, the other models optimized for the second annotation period achieved weaker performance in this case. The model with an extra 1,500 randomly selected comments from the second annotation achieved almost the same low F1-score as the base model without fine-tuning.

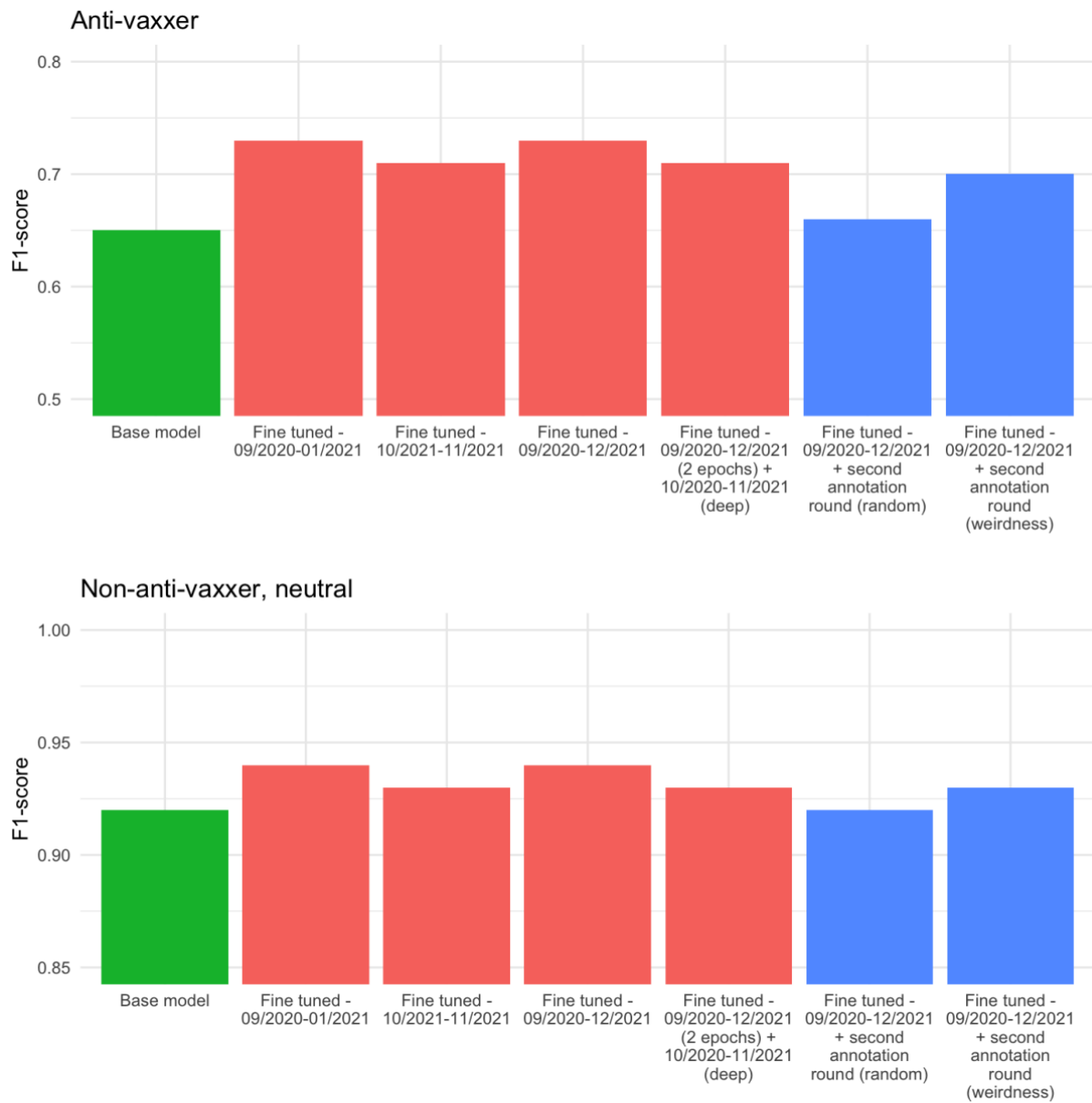

Figure 2. F1-score for models tested in the first period of annotation

Prior to the detailed analysis of classification performance, we examined how confident the models in the prediction were in the case of the test dataset coming from the second annotation period. To do this, we calculated a simple confidence value which was the absolute value of 0.5 minus the probability of anti-vaxxer classification given by the BERT model. The minimum value of this statistic is zero, and the maximum value is 0.5. A higher value means higher model confidence. We calculated this value for the test data and a subset of the test data with lower and higher weirdness values (see Table 1). As expected for all the models, the confidence value was low for comments with high weirdness and high for comments with low weirdness scores. Our base model had the lowest confidence value. The fine-tuned model had a higher confidence value, and the confidence value of the fine-tuned models and the models with the boosted classification samples did not differ significantly. Without knowing the true classification results based on model confidence, we would expect the same performance for the fine-tuned and the extended models. As we will see in the next section, the results did not confirm this expectation.

|  | Weird<0.9 | All test data | Weird>1.2 |
|---|---|---|---|
| Base model | 0.39 | 0.18 | 0.16 |
| Fine tuned - 09/2020-01/2021 | 0.44 | 0.28 | 0.25 |
| Fine tuned - 10/2021-11/2021 | 0.42 | 0.27 | 0.26 |
| Fine tuned - 09/2020-12/2021 | 0.43 | 0.31 | 0.27 |
| Fine tuned - 09/2020-12/2021 (2 epochs) + 10/2020-11/2021 (deep) | 0.44 | 0.28 | 0.26 |
| Fine tuned - 09/2020-12/2021 + second annotation round (random) | 0.41 | 0.27 | 0.26 |
| Fine tuned - 09/2020-12/2021 + second annotation round (weirdness) | 0.44 | 0.26 | 0.24 |

Table 1. Confidence level of models

Figure 3. presents the results of the comments used to test the classification quality for November 2021. There were 2,193 comments in the test database, none of which were used for fine-tuning or model classification. As expected, the overall quality of classification for these comments was much lower. The base model using annotated comments from the first period without fine-tuning only achieved an F1-score of 0.5 for anti-vaxxers and 0.66 for the rest of the comments. Fine-tuning without new training data did not improve the model quality for non-anti-vaxxers and neutral comments; in some cases, it even decreased it. For the anti-vaxxer category, fine-tuning boosted the F1 score. The model with the full unlabeled data and the model with the combination of swallow and deep learning had the same level of performance. The last two sets of models used classified comments from the time period of the test data. The performance of these two models was much higher than the earlier ones, so it seems that reclassification is inevitable for a changing corpus. The random sample from the annotated comments performed better based on the test data, achieving higher F1-scores both for the anti-vaxxer and the other class, so the special sampling design for the training dataset did not help the overall classification.

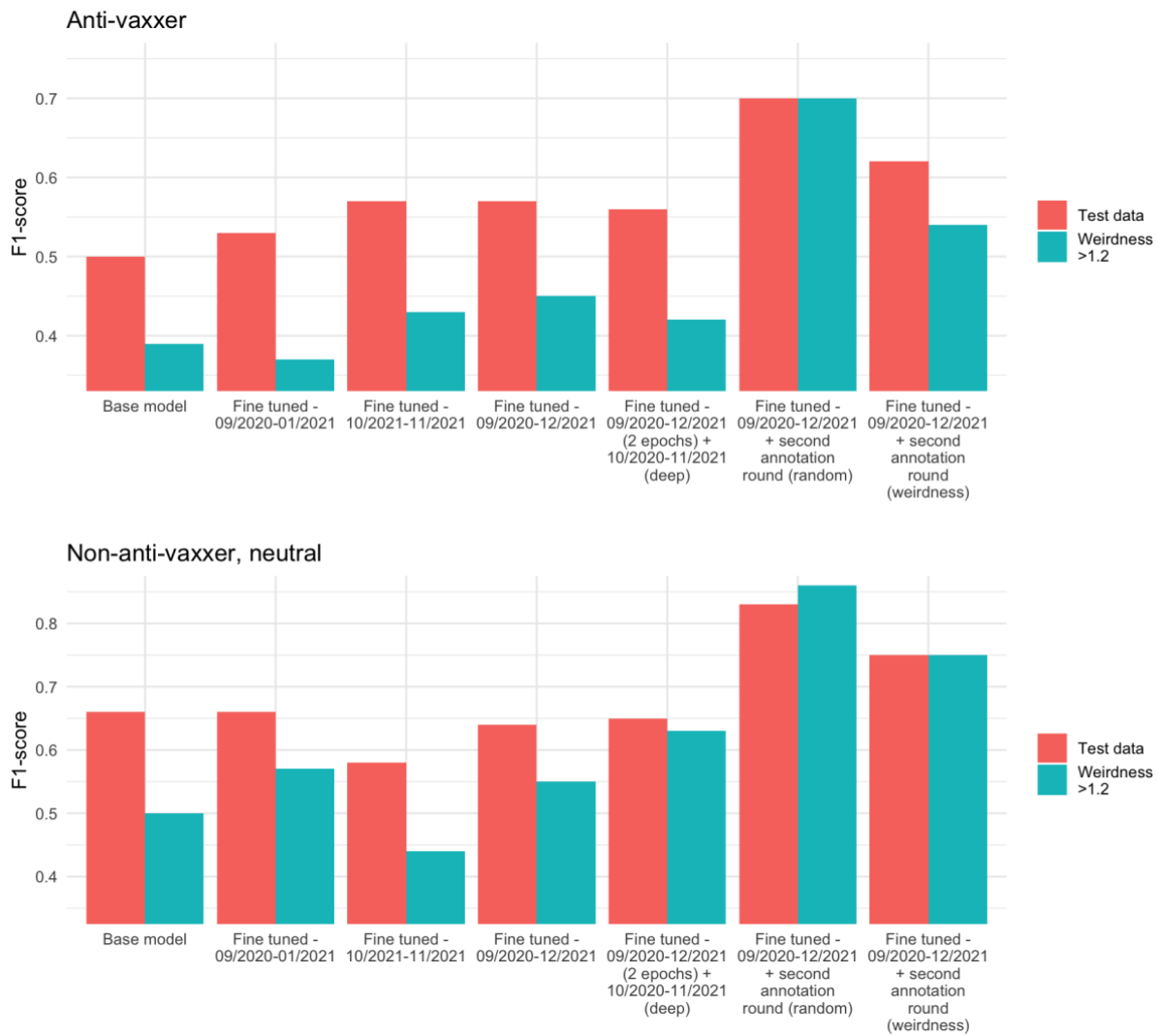

Figure 3. F1-score for models tested in the second period of annotation

In the next step, we checked how the classification model worked for the comments where the weirdness value was above 1.2 (see Figure 3.). About 10 percent of the annotated comments were above this threshold. The test data contained 223 comments that were above the 1.2 weirdness score. As expected for most models, the classification of these comments was more challenging than that of the "normal" comments; the model performance decreased sharply. There was only one exception, namely the extended model with the random sample from the second annotation. In contrast, the extended model with the sample based on weirdness probability had a weaker F1-score for the anti-vaxxers.

**Discussion**

Our results confirm the general result found in the literature on fine-tuning. Domain-based fine-tuning of transformer models can significantly improve their performance. Considering the temporal aspect of fine-tuning, the results clearly show that the best solution is to use all available unlabeled annotations in tuning the models (RQ1). Presumably, the more/better principle that is generally observed in larger language models applies here.

However, fine-tuning does not protect against model performance loss, but at least slows down the performance loss. In a rapidly changing linguistic environment, it is not possible to maintain model performance without adding new annotations. The timing of re-annotation is not generalizable, but indicators such as the weirdness index can help to shed light on corpus change dynamics. An important result of our analysis is that it is not advisable to select those comments that were less likely encountered by our model previously for new comment annotations. A more efficient solution is to randomly select the comments from the new period. The random sampling approach worked better than the specific weirdness-based sampling approach, even for classifying comments with high weirdness values. In the latter case, we saw that the model learned much worse on specific comments than on a more general training set (RQ2).

It is also a critical result that it is difficult to say anything about the performance of the models without repeated annotation. The confidence of the BERT models we tested was similar for both the fine-tuned and the extended models, although the latter performed better in reality. This model confidence can easily mislead researchers when no re-annotated data is available.

The creation of gold-standard categorizations will continue to be very important. However, these categorizations should not only always be domain-specific, but also need to take into account the temporal aspect. High performance of transformer models based on reinforced learning can only be maintained if a regular reclassification phase is built into the classification process, where the model is adapted to the new incoming data. Without this, our language models can quickly lose their predictive power.

**References**

Agarwal, O., & Nenkova, A. (2022). Temporal effects on pre-trained models for language processing tasks. *Transactions of the Association for Computational Linguistics*, *10*, 904-921.

Araci, D. (2019). Finbert: Financial sentiment analysis with pre-trained language models. *arXiv preprint arXiv:1908.10063*.

Basile, V. (2020). Domain adaptation for text classification with weird embeddings. In *Proceedings of the Seventh Italian Conference on Computational Linguistics* (pp. 1-7). CEUR-ws.

Beltagy, I., Lo, K., & Cohan, A. (2019). SciBERT: A pretrained language model for scientific text. *arXiv preprint arXiv:1903.10676*.

Chen, S., Neves, L., & Solorio, T. (2021). Mitigating Temporal-Drift: A Simple Approach to Keep NER Models Crisp. *arXiv preprint arXiv:2104.09742*.

Devlin, J., Chang, M. W., Lee, K., & Toutanova, K. (2018). Bert: Pre-training of deep bidirectional transformers for language understanding. *arXiv preprint arXiv:1810.04805*.

Florio, K., Basile, V., Polignano, M., Basile, P., & Patti, V. (2020). Time of your hate: The challenge of time in hate speech detection on social media. *Applied Sciences*, *10*(12), 4180.

Goel, R., Soni, S., Goyal, N., Paparrizos, J., Wallach, H., Diaz, F., & Eisenstein, J. (2016, November). The social dynamics of language change in online networks. In *International conference on social informatics* (pp. 41-57). Springer, Cham.

Huang, X., & Paul, M. (2018, July). Examining temporality in document classification. In *Proceedings of the 56th Annual Meeting of the Association for Computational Linguistics (Volume 2: Short Papers)* (pp. 694-699).

Huang, X., & Paul, M. (2019, July). Neural temporality adaptation for document classification: Diachronic word embeddings and domain adaptation models. In *Proceedings of the 57th Annual Meeting of the Association for Computational Linguistics* (pp. 4113-4123).

Koh, P. W., Sagawa, S., Marklund, H., Xie, S. M., Zhang, M., Balsubramani, A., ... & Liang, P. (2021, July). Wilds: A benchmark of in-the-wild distribution shifts. In *International Conference on Machine Learning* (pp. 5637-5664). PMLR.

Kulkarni, V., Al-Rfou, R., Perozzi, B., & Skiena, S. (2015, May). Statistically significant detection of linguistic change. In *Proceedings of the 24th international conference on world wide web* (pp. 625-635).

Larson, S., Singh, N., Maheshwari, S., Stewart, S., & Krishnaswamy, U. (2021, September). Exploring Out-of-Distribution Generalization in Text Classifiers Trained on Tobacco-3482 and RVL-CDIP. In *International Conference on Document Analysis and Recognition* (pp. 416-423). Springer, Cham.

Lazaridou, A., Kuncoro, A., Gribovskaya, E., Agrawal, D., Liska, A., Terzi, T., ... & Blunsom, P. (2021). Mind the gap: Assessing temporal generalization in neural language models. *Advances in Neural Information Processing Systems*, *34*, 29348-29363.

Lazer, D., Kennedy, R., King, G., & Vespignani, A. (2014). The parable of Google Flu: traps in big data analysis. *science*, *343*(6176), 1203-1205.

Lee, J., Yoon, W., Kim, S., Kim, D., Kim, S., So, C. H., & Kang, J. (2020). BioBERT: a pre-trained biomedical language representation model for biomedical text mining. *Bioinformatics*, *36*(4), 1234-1240.

Lukes, J., & Søgaard, A. (2018, October). Sentiment analysis under temporal shift. In *Proceedings of the 9th workshop on computational approaches to subjectivity, sentiment and social media analysis* (pp. 65-71).

Luu, K., Khashabi, D., Gururangan, S., Mandyam, K., & Smith, N. A. (2021). Time waits for no one! analysis and challenges of temporal misalignment. *arXiv preprint arXiv:2111.07408*.

Ma, X., Xu, P., Wang, Z., Nallapati, R., & Xiang, B. (2019, November). Domain adaptation with BERT-based domain classification and data selection. In *Proceedings of the 2nd Workshop on Deep Learning Approaches for Low-Resource NLP (DeepLo 2019)* (pp. 76-83).

Nemeskey, D. M. (2021). Introducing huBERT. In G. Berend, G. Gosztolya & V. Vincze (Eds.), *XVII. Magyar Számítógépes Nyelvészeti Konferencia* (pp. 3-14). JATEPress.

Ramponi, A., & Plank, B. (2020). Neural unsupervised domain adaptation in NLP---a survey. *arXiv preprint arXiv:2006.00632*.

Rijhwani, S., & Preoţiuc-Pietro, D. (2020, July). Temporally-informed analysis of named entity recognition. In *Proceedings of the 58th Annual Meeting of the Association for Computational Linguistics* (pp. 7605-7617).

Röttger, P., & Pierrehumbert, J. B. (2021). Temporal adaptation of BERT and performance on downstream document classification: Insights from social media. *arXiv preprint arXiv:2104.08116*.

Vaswani, A., Shazeer, N., Parmar, N., Uszkoreit, J., Jones, L., Gomez, A. N., ... & Polosukhin, I. (2017). Attention is all you need. *Advances in neural information processing systems*, *30*.

Wang, J., Lan, C., Liu, C., Ouyang, Y., Qin, T., Lu, W., ... & Yu, P. (2022). Generalizing to unseen domains: A survey on domain generalization. *IEEE Transactions on Knowledge and Data Engineering*.

Zsibrita, J., Vincze, V., & Farkas, R. (2013). magyarlanc: A Toolkit for Morphological and Dependency Parsing of Hungarian. In R. Mitkov, G. Angelova & K. Bontcheva (Eds.), *Proceedings of the International Conference Recent Advances in Natural Language Processing (RANLP 2013)* (pp. 763–771). Incoma Ltd.

*Funding*

The research was supported by the European Union within the framework of the RRF-2.3.1-21-2022-00004 Artificial Intelligence National Laboratory Program.

*Conflicts of Interest*

The authors declare no conflict of interest.

*Data availability*

Raw data is not available because of copyright issues, but access can be asked from the authors.

*Ethics approval*

As we did not include human subject in our research no ethical approval was needed.

*Author Contributions*

Conceptualization, Z.K. and B.K.; methodology, Z.K. K.B and B.K.; formal analysis, Z.K. K.B and B.K.; data curation, K.B.; writing—original draft preparation, Z.K. and B.K.; writing—review and editing, Z.K. and B.K; visualization, Z.K.; funding acquisition, Z.K.

## Supplementary

**Detailed description of the annotation**

Originally in the project, the annotators coded the comments into three categories. If the comment was clearly against vaccination by showing an intent not to get vaccinated, we have categorized it as anti-vaxxer content. Partially anti-vaxxer contents were more challenging to grab; they are the ones who do not reject vaccination at all but have specific concerns in different matters. Comments expressing uncertainty toward vaccination due to the lack of transparency and necessary information regarding vaccination were categorized as partially anti-vaxxer content. Moreover, discussions on unpredictability regarding the efficacy or the side effects of the vaccine among different age groups or the ones criticizing the vaccination process and the whole establishment in this context, which have an explicitly expressed effect on vaccination attitudes, were also categorized as partially anti-vaxxer comments. Comments expressing cherry-picking attitudes by choosing certain vaccines over other types - typically promoting 'western' vaccines against 'eastern' products - were also put into the partial anti-vaxxer category. The third category contains comments showing neutral and pro-vaccination attitudes.

However, seemingly non-relevant comments were added to this category since all the comments linked to articles were collected, even if the text is not about vaccination.

The annotator agreement was relatively low for this three-category classification. In the case of the first annotation period, the kappa value for the inter-coder agreement was 0.66. In the second annotation period, this kappa went down to 0.53. The initial transformer model performance was also low, so we decided to join the two anti-vaxxer categories. This paper runs all our analyses built on the two-category classification.

| Hungarian | English |
|---|---|
| olt | vaccinate |
| vakcina | vaccine |
| ad | give |
| kap | get |
| orosz | russian |
| kína | chinese |
| astra | astra |
| pfizer | pfizer |
| moderna | moderna |
| Szputnyik | sputnik |
| sem | nor |
| nem | not |
| % | % |
| megbízható | trusted |
| hatásos | effective |
| veszélyes | dangerous |

Table S1. – Keywords for filtering anti-vax comments

| Model | Class | Accuracy | f1-score | Precision | Recall |
|---|---|---|---|---|---|
| Base model | Anti-vaxxer |  | 0,65 | 0,66 | 0,64 |
| | Non anti-vaxxer, neutral | 0,87 | 0,92 | 0,92 | 0,92 |
| Fine tuned - 09/2020-01/2021 | Anti-vaxxer |  | 0,73 | 0,71 | 0,74 |
| | Non anti-vaxxer, neutral | 0,9 | 0,94 | 0,94 | 0,93 |
| Fine tuned - 10/2021-11/2021 | Anti-vaxxer |  | 0,71 | 0,78 | 0,65 |
| | Non anti-vaxxer, neutral | 0,89 | 0,93 | 0,91 | 0,95 |
| Fine tuned - 09/2020-12/2021 | Anti-vaxxer |  | 0,73 | 0,72 | 0,74 |
| | Non anti-vaxxer, neutral | 0,9 | 0,94 | 0,94 | 0,94 |
| | Anti-vaxxer | 0,89 | 0,71 | 0,69 | 0,73 |
| Fine tuned - 09/2020-12/2021 (2 epochs) + 10/2020-11/2021 (deep) | Non anti-vaxxer, neutral | | 0,93 | 0,94 | 0,93 |
| Fine tuned - 09/2020-12/2021 + second annotation round (random) | Anti-vaxxer | | 0,66 | 0,62 | 0,71 |
| | Non anti-vaxxer, neutral | 0,865 | 0,92 | 0,93 | 0,9 |
| Fine tuned - 09/2020-12/2021 + second annotation round (weirdness) | Anti-vaxxer | | 0,7 | 0,66 | 0,74 |
| | Non anti-vaxxer, neutral | 0,89 | 0,93 | 0,92 | 0,93 |

Table S2. Model performances for the first annotation round (09/2020-01/2021)

| Model | Class | Accuracy | f1-score | Precision | Recall |
|---|---|---|---|---|---|
| Base model | Anti-vaxxer | | 0.50 | 0.59 | 0.44 |
| | Non anti-vaxxer, neutral | 0.60 | 0.66 | 0.60 | 0.73 |
| Fine tuned - 09/2020-01/2021 | Anti-vaxxer | | 0.53 | 0.64 | 0.46 |
| | Non anti-vaxxer, neutral | 0.61 | 0.66 | 0.59 | 0.75 |
| Fine tuned - 10/2021-11/2021 | Anti-vaxxer | | 0.57 | 0.80 | 0.44 |
| | Non anti-vaxxer, neutral | 0.57 | 0.58 | 0.45 | 0.80 |
| Fine tuned - 09/2020-12/2021 | Anti-vaxxer | | 0.57 | 0.73 | 0.46 |
| | Non anti-vaxxer, neutral | 0.61 | 0.64 | 0.54 | 0.79 |
| Fine tuned - 09/2020-12/2021 (2 epochs) + 10/2020-11/2021 (deep) | Anti-vaxxer | | 0.56 | 0.69 | 0.46 |
| | Non anti-vaxxer, neutral | 0.61 | 0.65 | 0.57 | 0.77 |
| Fine tuned - 09/2020-12/2021 + second annotation round (random) | Anti-vaxxer | | 0.70 | 0.73 | 0.68 |
| | Non anti-vaxxer, neutral | 0.78 | 0.83 | 0.81 | 0.85 |
| Fine tuned - 09/2020-12/2021 + second annotation round (weirdness) | Anti-vaxxer | | 0.62 | 0.69 | 0.55 |
| | Non anti-vaxxer, neutral | 0.70 | 0.75 | 0.70 | 0.8 |

Table S3. Model performances for the second annotation round – test data

| Model | Class | Accuracy | f1-score | Precision | Recall |
|---|---|---|---|---|---|
| Base model | Anti-vaxxer | | 0.39 | 0.57 | 0.30 |
| | Non anti-vaxxer, neutral | 0.45 | 0.50 | 0.40 | 0.67 |
| Fine tuned - 09/2020-01/2021 | Anti-vaxxer | | 0.37 | 0.49 | 0.30 |
| | Non anti-vaxxer, neutral | 0.49 | 0.57 | 0.49 | 0.68 |
| Fine tuned - 10/2021-11/2021 | Anti-vaxxer | | 0.43 | 0.70 | 0.32 |
| | Non anti-vaxxer, neutral | 0.44 | 0.44 | 0.32 | 0.70 |
| Fine tuned - 09/2020-12/2021 | Anti-vaxxer | | 0.45 | 0.65 | 0.34 |
| | Non anti-vaxxer, neutral | 0.50 | 0.55 | 0.44 | 0.74 |
| Fine tuned - 09/2020-12/2021 (2 epochs) + 10/2020-11/2021 (deep) | Anti-vaxxer | | 0.42 | 0.54 | 0.35 |
| | Non anti-vaxxer, neutral | 0.55 | 0.63 | 0.55 | 0.73 |
| Fine tuned - 09/2020-12/2021 + second annotation round (random) | Anti-vaxxer | | 0.70 | 0.71 | 0.69 |
| | Non anti-vaxxer, neutral | 0.81 | 0.86 | 0.86 | 0.87 |
| Fine tuned - 09/2020-12/2021 + second annotation round (weirdness) | Anti-vaxxer | | 0.54 | 0.61 | 0.48 |
| | Non anti-vaxxer, neutral | 0.68 | 0.75 | 0.71 | 0.80 |

Table S4. Model performances for the second annotation round – test data, weirdness value greater than 1.2